\documentclass[final,3p,times]{elsarticle}

\usepackage{amssymb}
\usepackage{comment}
\usepackage{color}

\usepackage{ascmac}

\usepackage{algorithm}
\usepackage{algorithmic}
\usepackage{framed}

\usepackage{amsthm,amsmath,bm}

\usepackage{multirow}
\usepackage{graphicx}
\usepackage{color,colortbl}

\journal{arXiv}

\begin{document}

\begin{frontmatter}

\title{Policy Gradient Reinforcement Learning for Policy Represented by Fuzzy Rules: Application to Simulations of Speed Control of an Automobile}

\author{Seiji Ishihara}
\ead{ishihara\_s@mail.dendai.ac.jp}
\address{School of Science and Engineering, Tokyo Denki University\\Ishizaka, Hatoyama-machi, Hiki-gun, Saitama 350-0394, Japan}

\author{Harukazu Igarashi}
\address{College of Engineering, Shibaura Institute of Technology\\3-7-5, Toyosu, Koto-ku, Tokyo 135-8548, Japan}

\begin{abstract}
A method of a fusion of fuzzy inference and policy gradient reinforcement learning has been proposed that directly learns, as maximizes the expected value of the reward per episode, parameters in a policy function represented by fuzzy rules with weights. A study has applied this method to a task of speed control of an automobile and has obtained correct policies, some of which control speed of the automobile appropriately but many others generate inappropriate vibration of speed. In general, the policy is not desirable that causes sudden time change or vibration in the output value, and there would be many cases where the policy giving smooth time change in the output value is desirable. In this paper, we propose a fusion method using the objective function, that introduces defuzzification with the center of gravity model weighted stochastically and a constraint term for smoothness of time change, as an improvement measure in order to suppress sudden change of the output value of the fuzzy controller. Then we show the learning rule in the fusion, and also consider the effect by reward functions on the fluctuation of the output value. As experimental results of an application of our method on speed control of an automobile, it was confirmed that the proposed method has the effect of suppressing the undesirable fluctuation in time-series of the output value. Moreover, it was also showed that the difference between reward functions might adversely affect the results of learning.
\end{abstract}

\begin{keyword}
fuzzy controller, reinforcement learning, policy gradient method, center of gravity model, speed control problem

\end{keyword}

\end{frontmatter}


\section{Introduction}
Deep learning has made remarkable progress in recent years. It has been applied to the action control of robots \cite{Duan} and game agents in classical action video games \cite{Mnih} and intellectually challenging games such as Go \cite{Silver}. However, a large amount of learning data and computational resources are required for large neural network models to learn data precisely \cite{Silver}.

A difficult analysis of weights in deep neural networks is also necessary to understand the properties of a control system after learning, i.e., which features are important as input data and which control rules are obtained within the neural networks by learning. Fuzzy controllers are based on if-then rules, in which premises and conclusions are expressed using natural languages. That makes reasoning processes very readable and allows system designers to insert very easily a priori knowledge of human experts into the rules. Moreover, fuzzy control systems do not need such huge amounts of computation costs as deep learning does.

Many researchers have worked to find automatic methods that allow self-tuning of fuzzy control systems by reinforcement learning. This approach can combine the benefits of various fuzzy control systems with reinforcement learning, since the parameters included in the membership functions of fuzzy control rules can also be learned by reinforcement learning, even if there is no teacher data in the input and output of the control system. For reinforcement learning, fuzzy rules are appropriate for building control systems that can deal with continuous and layered system states \cite{Berenji} other than the readability of inference rules.

In those studies, Igarashi and Ishihara proposed an integrated method, where a control policy was expressed by fuzzy rules and parameters in the rules were learned by a policy gradient reinforcement algorithm \cite{Igarashi2}. They applied this method to a speed-control problem for an automobile and found some undesirable solutions with high-frequency oscillation in the running speed of an automobile. In this paper, we propose a way to improve that method using an output decision based on a gravity-center model and a time-smoothing constraint. We also derive learning rules for the weight parameters of fuzzy control rules and show the importance of designing the reward functions given to this example problem.

This paper is organized as follows: Section 2 summarizes related work on combining fuzzy control systems and reinforcement leaning. Details of the integrated method from the previous work \cite{Igarashi2} are described in Section 3, which is the basis of our control system. Section 4 describes the gravity-center model and a time-smoothing constraint term added to the objective function used in the control policy. New learning rules for the weight parameters included in fuzzy rules are applied in the modified controlling system. Section 5 presents a case study of controlling the speed of an automobile. The results of learning experiments and a discussion are given in Section 6. Section 7 provides a summary of this paper and touches on our future work. The derivation of learning rules in the new model proposed in Section 4 is shown in the Appendix. 

\section{Related Work} \label{sec2}

\subsection{Fuzzy control and reinforcement learning}

Research on combining fuzzy control and reinforcement learning can be roughly divided into two categories depending on the type of reinforcement learning. The first type uses value-based reinforcement learning with an assumption of Markov Decision Processes (MDPs) on the environments and the policies of agents \cite{Jouffe,Oh,Horiuchi,Hoshino}. The second type uses policy-based reinforcement learning called a policy gradient method, which learns the parameters in a policy without calculating value functions \cite{Wang,Igarashi2}.

The fuzzy rules used in the first type of control system usually describe a state of the system in their antecedent parts. The parameter values \cite{Jouffe,Oh} or functions \cite{Horiuchi} in their consequent parts describe Q values of actions. However, fuzzy sets were not used to describe the output variables in the consequent parts. The control system calculated Q values only for the discrete actions of agents to be controlled. Moreover, there were no weight parameters representing the confidence or importance of the rules to reinforce right and necessary rules and to suppress wrong and unnecessary rules. 

The second type of control system used policy-based reinforcement learning, called a policy gradient method. That learning method originates from Williams' REINFORCE algorithm \cite{Williams}, and it is an approach that computes policy gradient vectors with respect to parameters in the policy function and improves the policy by adjusting the parameters in the gradient direction \cite{Williams,Baxter,Igarashi1}. An integrated method was proposed by Wang et al. \cite{Wang} using a policy gradient method called the GPOMDP algorithm, which was proposed by Baxter and Bartlett \cite{Baxter}. However, agent actions were restricted to being discrete. Fuzzy sets were used only in antecedent parts and fixed without being learned. There were no weight parameters of rules. 
 
To compensate for those imperfections, we proposed the integrated method of fuzzy control and reinforcement learning based on the policy gradient algorithm \cite{Igarashi2}. Our method allows fuzzy sets in both the antecedent and the consequent parts of system control rules, and it can learn the membership functions and weight parameters of the fuzzy rules to maximize the expected return per episode locally. Therefore, system designers can design fuzzy control rules more freely than with other combining systems. 

\subsection{Case studies of applying the integrated method to control problems}

Sugimoto et al. \cite{Sugimoto} applied the integrated method proposed by Igarashi and Ishihara \cite{Igarashi2} to the action decision of a soccer robot in the RoboCup Small Size League \cite{RoboCup1,RoboCup2}. The robot control system learns a stochastic policy of human decision making when a robot holds a ball. This is an application of an integrated method to supervised learning problems. However, the robot is restricted to three actions: shot, pass and dribble. The control system simply makes a decision on which kind of actions to take from these three categories in static scenes chosen from the robot soccer games\footnote{Strength to choose each action is included in the consequent part of each rule.}. They succeeded in learning a stochastic policy of human decision making in 25 of the 30 training scenes. However, these studies did not deal with cases where a real robot takes real actions, changes its environments, and utilizes the rewards given for the results of the actions made to improve its policy.

Tsuchiya et al. applied the integrated method of Igarashi and Ishihara \cite{Igarashi2} to reinforcement learning in fuzzy control problems \cite{Tsuchiya}. In that paper, the control system learned rule weights of fuzzy rules for controlling a vehicle's speed to keep a particular distance from another vehicle moving in front of it. The output variables of the control system were continuous variables corresponding to the intensities of accelerating and braking. The fuzzy controller with the learned weights controlled a vehicle's speed in the same manner as, or more quickly than, humans selected rules and the standard min-max-gravity method\footnote{Humans select appropriate fuzzy rules from among the twenty rules, and inference results are combined using min and max operations.}. However, their experiments showed that the integrated method easily gives undesirable solutions that cause high-frequency oscillation in the car's running speed\footnote{High-frequency oscillation in the speed of a car brings a feeling of discomfort, and such fine-grained control is not easy in practice. An example of high-frequency oscillation is shown in Fig. 5 of the earlier work \cite{Tsuchiya}, and another one is shown in Fig. \ref{fig:3}(a) of this paper.}. They concluded that the control system needs some regularization in policy and that the reward functions should be modified to avoid such undesirable solutions. 

Policies that output values with rapid changes and oscillation in time are not desirable. In this paper, we propose a regularization of the objective function, using the gravity-center model and adding the time-smoothing constraint term to the objective function, while designing a new reward function to avoid the undesirable solutions.

\section{Summary of base theory} \label{sec3}

We improve the learning method of the fuzzy control system proposed by Igarashi and Ishihara \cite{Igarashi2} as used in the work of Tsuchiya et al. \cite{Tsuchiya}. Before describing the details, we summarize the original base method \cite{Igarashi2}. The fuzzy control rules, the objective function and policy, and the learning rules of the rule weights are explained in Sections 3.1, 3.2 and 3.3, respectively. The learning rules of membership functions are not mentioned here because they are outside the scope of this paper.

\subsection{Fuzzy control rules}

In the previous work \cite{Igarashi2}, the following fuzzy rules were used to control an agent's behavior,
\begin{eqnarray}
\textrm{Rule}\ i :\ \textrm{if}\ (x_1\ \textrm{is}\ A^{i}_1)\ \textrm{and}\ \dots\ \textrm{and}\ (x_M\ \textrm{is}\ A^{i}_M)\ \textrm{then}\ (y_1\ \textrm{is}\ B^{i}_1)\ \textrm{and}\ \dots\ \textrm{and}\ (y_N\ \textrm{is}\ B^{i}_N)\ \textrm{with}\ \theta_i. 
\label{eq:1}
\end{eqnarray}
Subscript $i$ ($=1,2,\cdots,n_{\textrm{R}}$) is a rule number, $x_j$ ($j=1,2,\cdots,M$) are input variables describing a state. $y_k$ ($k=1,2,\cdots,N$) are output variables describing actions for an agent to take. $A^{i}_j$/$B^{i}_k$ is a linguistic fuzzy expression for $x_j$/$y_k$ included in the $j$-th antecedent/$k$-th consequent parts of the $i$-th rule. $A^{i}_j(x_j)$ and $B^{i}_k(y_k)$ are membership functions, and $\theta_i$ ($\geq 0$) is a weight parameter of the $i$-th rule. All rule weights {$\theta_i$} are parameters to be learned.

\subsection{Objective function and policy}

Assume that an agent selects an action at every time step to minimize an objective function defined by
\begin{eqnarray}
E(y;x,\theta,A,B)=-\sum_{i=1}^{n_{\textrm{R}}}\theta_iA^i(x)B^i(y).
\label{eq:2}
\end{eqnarray}
The membership degree $A^i(x)$/$B^i(y)$ of input $x$/output $y$ in the antecedent/consequent part in Rule $i$ is defined by
\begin{eqnarray}
A^i(x)\equiv \prod_{j=1}^M A^i_j(x_j),
\label{eq:3}
\end{eqnarray}
\begin{eqnarray}
B^i(y)\equiv \prod_{k=1}^N B^i_k(y_k).
\label{eq:4}
\end{eqnarray}
The product in Eq. \eqref{eq:3}/\eqref{eq:4} means that the truth value of the antecedent/consequent part is calculated by the product of the degrees of inclusion in the fuzzy sets $A^i_j$/$B^i_k$ of input/output variables $x_j$/$y_k$. The objective function in Eq. \eqref{eq:2} indicates how much all rules support output value $y$ when $x$ is input to the control system.

The control system determines output $y$ for input $x$ at every time step $t$. The policy is defined by a mapping function from input state $x(t)$ to action $y(t)$. Let the policy have a stochastic character and be defined by the following Boltzmann distribution function:
\begin{eqnarray}
\pi (y(t);x(t),\theta,A,B) &\equiv& \frac{e^{-E(y(t);x(t),\theta,A,B)/T}}{\sum_{y}e^{-E(y;x(t),\theta,A,B)/T}},
\label{eq:5}
\end{eqnarray}
where $E(y;x)$ is called an objective function or energy function. Igarashi and Ishihara used Eq. \eqref{eq:2} as the objective function in Eq. \eqref{eq:5}. Parameter $T$ is called temperature. It controls the intensity of the expectation of $E(y)$ with respect to distribution $\pi(y)$ and the randomness of the action decision.

\subsection{Learning rules}
A learning rule of weight parameter $\theta_i$ is given \cite{Williams,Igarashi1} as 
\begin{eqnarray}
\Delta\theta_i &=& \varepsilon r\sum_{t=0}^{L-1}e_{\theta_i}(t),
\label{eq:pre9}
\end{eqnarray}
where $L$ is episode length, $r$ is a reward function given to an episode, and $\varepsilon$ ($>0$) is a learning rate. $e_{\theta_i}(t)$ is characteristic eligibility defined by $e_{\theta_i} \equiv {\partial\textrm{ln}\pi}/{\partial \theta_i}$ at time $t$ and is shown in the previous work \cite{Igarashi2} as
\begin{eqnarray}
e_{\theta_i}(t) = \frac{1}{T}A^{i}(x(t))\left\{B^{i}(y(t))-\left\langle B^{i}(y)\right\rangle_{\pi(y;x(t))}\right\},
\label{eq:pre10}
\end{eqnarray}
where
\begin{eqnarray}
\left\langle B^{i}(y)\right\rangle_{\pi(y;x(t))} \equiv \sum_{y}B^{i}(y)\ \pi(y;x(t),\theta,A,B).
\label{eq:pre11}
\end{eqnarray}

\section{Modification of Inference System and Learning Method} \label{sec4}

\subsection{Defuzzying based on a gravity-center model}

Tsuchiya et al. proposed a defuzzying algorithm based on a gravity-center model \cite{Tsuchiya}, where the control system calculates the gravity-center of $y$ and decides the value as a system output.
Mizumoto proposed a fuzzy inference model called the ``product-sum-gravity method'' \cite{Mizumoto}, in which min-max operations are replaced with algebraic product and sum operations as done by Igarashi and Ishihara \cite{Igarashi2}. 
The two models are similar to each other. However, the gravity-center $y_g$ of algebraic sum $B(y)=\sum_kB_k(y)$, i.e.,
\begin{eqnarray}
y_{\textrm{g}} &\equiv& \frac{\int y B(y)dy}{\int B(y)dy}\
\label{eq:13}
\end{eqnarray}
was used in Mizumoto's model. In Tsuchiya's model, another gravity-center $y_G$ was used. That model includes rule weight $\theta$ and is defined by the expectation of $y$ with respect to $\pi$ as 
\begin{eqnarray}
y_{\textrm{G}}(t) &\equiv& \sum_yy\cdot\pi (y;x(t),\theta,A,B)\nonumber\\
&=& \left\langle y \right\rangle_{\pi(y;x(t))}.
\label{eq:6}
\end{eqnarray}

Tsuchiya's control system chooses a discrete value as output $y(t)$ at time $t$ according to the probabilistic distribution function in Eq. \eqref{eq:5} in the learning phase. The gravity-center $y_{\textrm{G}}$ was not applied to the learning phase, while it was used deterministically as the output of the control system in the evaluation phase of $\theta$ after learning. We propose utilizing $y_{\textrm{G}}$ also in the learning phase as described in the next section.

\subsection{Learning phase using gravity-center model and time-smoothing constraint}

We propose an objective function $E'(y)$ to minimize the squared error between $y_{\textrm{G}}(t)$ in Eq. \eqref{eq:6} and output $y$ under a time-smoothing constraint defined as
\begin{eqnarray}
E'(y) &=& \frac{1}{2}(y-y_{\textrm{G}}(t))^2+\lambda (y-y(t-1))^2.
\label{eq:7}
\end{eqnarray}
The first and second terms in Eq. \eqref{eq:7} mean that output $y$ at time $t$ should be near $y_{\textrm{G}}(t)$ and $y(t-1)$ to avoid a large change in $y$. Parameter $\lambda$ ($\geq 0$) represents the strength of the second term.

Instead of the policy in Eq. \eqref{eq:5}, we propose a policy $\pi'$ with the new objective function $E'(y)$ as
\begin{eqnarray}
\pi' (y(t)) &\equiv& \frac{e^{-E'(y(t))/T'}}{\sum_{y}e^{-E'(y)/T'}}.
\label{eq:8}
\end{eqnarray}
The control system selects a discrete value as output $y(t)$ according to the stochastic policy $\pi'$ in the learning phase. In the evaluation phase of $\theta$, the system takes the minimal solution $y_{\textrm{O}}(t)$ of $E'(y)$ in Eq. \eqref{eq:7} as $y(t)$. Precisely, the solution is expressed as
\begin{eqnarray}
y_{\textrm{O}}(t) &\equiv& \frac{y_{\textrm{G}}(t)+2\lambda y(t-1)}{1+2\lambda}.
\label{eq:14}
\end{eqnarray}

\subsection{Learning rules of weights $\{\theta_i\}$ with policy $\pi'$}

The learning rule of $\theta_i$ when using policy $\pi'$ is 
\begin{eqnarray}
\Delta\theta_i &=& \varepsilon ' r\sum_{t=0}^{L-1}e'_{\theta_i}(t).
\label{eq:9}
\end{eqnarray}
This is the same rule as Eq. \eqref{eq:pre9}. Substituting Eqs. \eqref{eq:7} and \eqref{eq:8} into $e'_{\theta_i} \equiv {\partial\textrm{ln}\pi'}/{\partial \theta_i}$, $e'_{\theta_i}(t)$ is given by
\begin{eqnarray}
e'_{\theta_i}(t) = -\frac{A^{i}(x(t))}{TT'}\left\langle(y-y_{\textrm{G}}(t))B^{i}(y)\right\rangle_{\pi(y;x(t))}\left(y(t)-\left\langle y\right\rangle_{\pi'(y;x(t))}\right).
\label{eq:10}
\end{eqnarray}
All rule weights $\{\theta_i\}\ (i=1,2,\cdots,n_{\textrm{R}})$ are updated at the end of each episode. The meanings of Eqs. \eqref{eq:9} and \eqref{eq:10} are as follows: The degree of reward $r$ and membership degrees of the antecedent and consequent parts control the amount of update of $\theta_i$. The learning rule approaches output $y(t)$ to the gravity center $y_{\textrm{G}}(t)$. A derivation of Eq. \eqref{eq:10} is given in the Appendix.

\section{Case Studies} \label{sec5}

\subsection{Speed control of an automobile}

Tsuchiya et al. applied their method to cases of speed control of an automobile \cite{Tsuchiya}. This section summarizes the details of the control problems. Assume that two cars are running in the same direction on a one-dimensional straight-line road. The forward car is running at a constant speed. The system controls the speed of the rear running car. The goal condition required of the system is that the distance between two cars is adjusted to be in a specific range and maintained during a limited period of time.

\subsection{Fuzzy control rules}

\begin{table}[!tb]
\caption{Fuzzy control rules used in the speed control problems of an automobile. Here, ``ac.'' means acceleration and ``de.'' means deceleration.}
\label{table:1}
\begin{center}
\begin{tabular}{c|c|c|c}
\noalign{\hrule height 0.4mm}
\multirow{2}{*}{Rule} & \multirow{2}{*}{Following distance} & Speed of the & Operation of the\\
& & following car & following car\\
\hline
1 & \multirow{10}{*}{long} & \multirow{5}{*}{fast} & strong ac.\\
\cline{1-1}\cline{4-4}
2 & & & weak ac.\\
\cline{1-1}\cline{4-4}
3 & & & strong de.\\
\cline{1-1}\cline{4-4}
4 & & & weak de.\\
\cline{1-1}\cline{4-4}
5 & & & none\\
\cline{1-1}\cline{3-4}
6 & & \multirow{5}{*}{slow} & strong ac.\\
\cline{1-1}\cline{4-4}
7 & & & weak ac.\\
\cline{1-1}\cline{4-4}
8 & & & strong de.\\
\cline{1-1}\cline{4-4}
9 & & & weak de.\\
\cline{1-1}\cline{4-4}
10 & & & none\\
\hline
11 & \multirow{10}{*}{short} & \multirow{5}{*}{fast} & strong ac.\\
\cline{1-1}\cline{4-4}
12 & & & weak ac.\\
\cline{1-1}\cline{4-4}
13 & & & strong de.\\
\cline{1-1}\cline{4-4}
14 & & & weak de.\\
\cline{1-1}\cline{4-4}
15 & & & none\\
\cline{1-1}\cline{3-4}
16 & & \multirow{5}{*}{slow} & strong ac.\\
\cline{1-1}\cline{4-4}
17 & & & weak ac.\\
\cline{1-1}\cline{4-4}
18 & & & strong de.\\
\cline{1-1}\cline{4-4}
19 & & & weak de.\\
\cline{1-1}\cline{4-4}
20 & & & none\\
\noalign{\hrule height 0.4mm}
\end{tabular}
\end{center}
\end{table}

The $i$-th fuzzy rule used in a previous work \cite{Tsuchiya} to control the speed of an automobile is as follows:
\begin{eqnarray}
\textrm{Rule}\ i :\ \textrm{if}\ (x_1\ \textrm{is}\ A^{i}_1)\ \textrm{and}\ (x_2\ \textrm{is}\ A^{i}_2)\ \textrm{then}\ (y_1\ \textrm{is}\ B^{i}_1)\ \ \textrm{with}\ \theta_i.
\label{eq:12}
\end{eqnarray}
Input $x$ to the control system has two components $x_1$ and $x_2$ shown in the antecedent part of this rule. In this example, $x_1$ is the distance between two cars and $x_2$ is the speed of the following car. Output $y$ has only one component $y_1$ in the consequent part of rule \eqref{eq:12}. $y_1$ represents the degree of accelerating or braking by the following car. Membership functions $A^{i}_1$ and $A^{i}_2$ correspond to the concepts of ``long/short distance'' and ``fast/slow speed,'' respectively. Membership function $B^{i}_1$ takes one of the five representations such as ``strong/weak acceleration,'' ``strong/weak deceleration,'' and ``doing nothing''. Table \ref{table:1} shows the twenty fuzzy control rules given by all possible combinations of $A^{i}_1$, $A^{i}_2$ and $B^{i}_1$.

There are some rules that seem inappropriate intuitively. Rule no. 8 states that an agent in the following car must step on the brake pedal ``strongly'' when the distance to the leading car is ``long'' and the speed of the following car is ``slow.'' That rule prevents the following car from catching up with the leading car. Rule no. 11 states that the agent must step on the accelerator pedal ``strongly'' when the distance to the leading car is ``short'' and the speed of the following car is ``fast.'' That rule leads to rear-end collisions very easily. The objectives of the learning are to decrease the weight values of such inappropriate rules and to reinforce appropriate rules to obtain a policy that satisfies the control requirements.

\subsection{Membership Functions}

\begin{figure}[tb]
\centering
\begin{center}
\includegraphics[bb=0 0 341 258, scale=1]{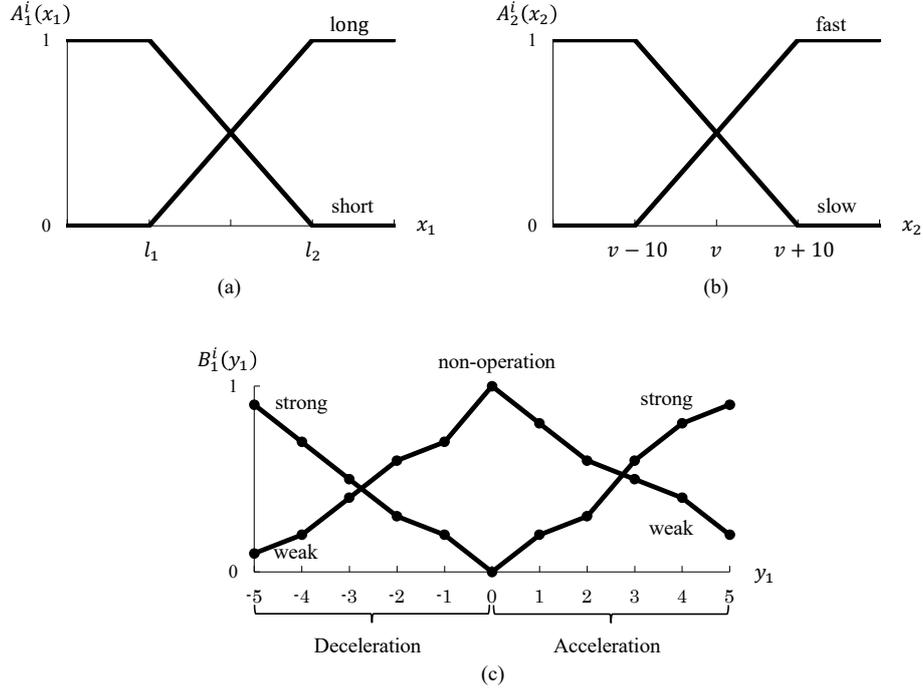}
\end{center}
\vspace{-5mm}
\caption{Membership functions. (a) Membership functions $A^{i}_1(x_1)$ of distance $x_1$ between two cars. (b) Membership functions $A^{i}_2(x_2)$ of the following car's speed $x_2$. (c) Membership functions $B^{i}_1(y_1)$ of the system's output $y_1$.}
\label{fig:1}
\end{figure}

Membership functions $A^{i}_1$, $A^{i}_2$ and $B^{i}_1$ in rule \eqref{eq:12} were set as shown in Fig.\,\ref{fig:1} of the previous work \cite{Tsuchiya}. Two membership functions on $A^{i}_1$ are defined in Fig.\,\ref{fig:1}(a) using polygonal lines and the required distance $[l_1, l_2]$. Two membership functions on $A^{i}_2$ are defined in Fig.\,\ref{fig:1}(b) using polygonal lines, and the leading car's speed $v$. The five functions on $B^{i}_1$ are defined in Fig.\,\ref{fig:1}(c). While almost all of the functions are continuous, there is one exception. ``Doing nothing'' defined in Fig.\,\ref{fig:1}(c) has a crisp membership function that takes one when $y_1=0$, or else takes zero.

Discrete time is set as $\Delta t=1$ [sec]. It is assumed that acceleration of $2y_1$ [km/h/s] is added to a car so that its speed reaches 100 [km/h] in 10 seconds when the system output $y_1$ continues to take its maximum value five. For deceleration, the same acceleration $2y_1$ is added to a car in the reverse direction of running.

\section{Computer Experiments} \label{sec6}

\begin{table}[!tb]
\caption{Objective function, output, learning rule and constraint condition of the three methods (i), (ii) and (iii).}
\label{table:5}
\begin{center}
\begin{tabular}{c|c|c|c|c|c}
\noalign{\hrule height 0.4mm}
\multirow{4}{*}{Meth.} & \multicolumn{3}{|c|}{\multirow{2}{*}{Learning phase}} & Evaluation & \\
& \multicolumn{3}{|c|}{} & phase & Constraint\\
\cline{2-5}
& Objective & \multirow{2}{*}{Output} & Learning & \multirow{2}{*}{Output} & condition\\
& function & & rule & & \\
\hline
\textrm{(i)} & $E(y)$ & $\pi(y)$ & (6) \& (7) & $y_{\textrm{G}}$ & ---\\
\hline
\textrm{(i\hspace{-.08em}i)} & \multirow{2}{*}{$E'(y)$} & \multirow{2}{*}{$\pi'(y)$} & \multirow{2}{*}{(14) \& (15)} & \multirow{2}{*}{$y_{\textrm{O}}$} & $\lambda = 0$ \\
\cline{1-1}\cline{6-6}
\textrm{(i\hspace{-.08em}i\hspace{-.08em}i)} & & & & & $\lambda = 0.06$ \\
\hline
\noalign{\hrule height 0.4mm}
\end{tabular}
\end{center}
\end{table}

We conducted experiments to test the three methods shown as (i), (ii) and (iii) in Table \ref{table:5}. Method (i) is that of Tsuchiya et al. \cite{Tsuchiya}. Our improvements proposed in this paper are included in (ii) and (iii). The time-smoothing constraint term in Eq. \eqref{eq:7} is not used in (ii). Output $y_1$ takes a discrete value as $y_1=-5.0+0.1h\ (h=0,1,\cdots ,100)$ in both the learning phase and the evaluation phase.

\subsection{Learning}

\begin{table}[!tb]
\caption{Initial conditions in 16 control problems.}
\label{table:2}
\begin{center}
\begin{tabular}{c|c|c|c|c|c}
\noalign{\hrule height 0.4mm}
\multirow{4}{*}{Problem} & \multicolumn{5}{|c}{Conditions}\\
\cline{2-6}
 & \multicolumn{2}{|c|}{Initial speed (\textrm{km/h})} & Initial & \multicolumn{2}{|c}{Target}\\
\cline{2-3}
 & \multirow{2}{*}{Forward car} & \multirow{2}{*}{Following car} & following & \multicolumn{2}{|c}{distance (\textrm{m})}\\
\cline{5-6}
 & & & distance (\textrm{m}) & \multicolumn{1}{|c}{\ \ \ $l_1$\ \ \ } & \multicolumn{1}{|c}{$l_2$}\\
\hline
1 & 20 & 30 & 50 & 30 & 45\\
\hline
2 & 20 & 30 & 50 & 10 & 15\\
\hline
3 & 20 & 30 & 10 & 30 & 45\\
\hline
4 & 20 & 30 & 10 & 10 & 15\\
\hline
5 & 30 & 30 & 50 & 30 & 45\\
\hline
6 & 30 & 30 & 50 & 10 & 15\\
\hline
7 & 30 & 30 & 10 & 30 & 45\\
\hline
8 & 30 & 30 & 10 & 10 & 15\\
\hline
9 & 50 & 30 & 50 & 30 & 45\\
\hline
10 & 50 & 30 & 50 & 10 & 15\\
\hline
11 & 50 & 30 & 10 & 30 & 45\\
\hline
12 & 50 & 30 & 10 & 10 & 15\\
\hline
13 & 60 & 30 & 50 & 30 & 45\\
\hline
14 & 60 & 30 & 50 & 10 & 15\\
\hline
15 & 60 & 30 & 10 & 30 & 45\\
\hline
16 & 60 & 30 & 10 & 10 & 15\\
\noalign{\hrule height 0.4mm}
\end{tabular}
\end{center}
\end{table}

Tsuchiya et al. prepared 16 problems for learning experiments. The initial speeds of the two cars, initial distances between them, and target distances $[l_1, l_2]$ for the 16 problems are shown in Table \ref{table:2}. Weight values $\{\theta_i\}\ (i=1,\cdots,20)$ given to the 20 rules in Table \ref{table:1} are learned by the three reinforcement learning methods (i), (ii) and (iii) in Table \ref{table:5}. 

Controlling the following car's speed starts at $t=0$ and ends at $t=110$. That means the length $L$ of every episode is 110. If the distance becomes shorter than zero, i.e., collision, or not shorter than 200 [m], the system stops controlling and the episode is terminated. The goal of controlling is satisfying the following two requisites\footnote{There is no requirement that the two requisites must be satisfied as soon as possible.} on the distance between two cars.
\begin{description}
\vspace{2mm}
  \item[Requisite 1:] The distance between two cars is in the specified range $[l_1, l_2]$.
  \item[Requisite 2:] The preceding requisite holds from $t=80$ to $t=110$. 
\vspace{2mm}
\end{description}

We prepared two types of reward functions. The first one, $r_1$, is a reward function that gives only penalties. The second one, $r_2$, gives rewards and penalties similar to those used by Tsuchiya et al. \cite{Tsuchiya}. The two reward functions $r_1$ and $r_2$ are defined in five cases as follows:
\begin{description}
\vspace{2mm}
  \item[Case 1 (Both Req. 1 and 2 satisfied):]If the following car enters the range $[l_1, l_2]$ at time $t_{\textrm{in}}$ that is equal to or earlier than $t=80$ and maintains the distance until $t=110$, then let $r_1=0$ and $r_2={0.01}/{(t_{\textrm{in}}+1)}$.
  \item[Case 2 (Only Req. 1 satisfied):]If the following car enters the range $[l_1, l_2]$ at time $t_{\textrm{in}}$ that is later than $t=80$ and maintains the distance until $t=110$, then let $r_1=(80-t_{\textrm{in}})/100+c$ and $r_2={0.01}/{(t_{\textrm{in}}+1)}$.
  \item[Case 3 (Neither Req. 1 nor 2 satisfied):]If the following car has not entered the range $[l_1, l_2]$ at $t=110$, then let \\$r_1=r_2=-|((l_1+l_2)/2-x_1)/20000|+c$D
  \item[Case 4 (Collision occurs):]If the distance becomes negative, that means the following car has collided with the leading car. In this case, let $r_1=r_2=-x_1^2/100+c$.
  \item[Case 5 (Aborted due to excessive distance)]If the distance becomes equal to or larger than 200 [m], then the control system quits controlling. Rewards are set as $r_1=r_2=(t_{\textrm{far}}-110)/100+c$, where $t_{\textrm{far}}$ is the time when the control system quits controlling.
\vspace{2mm}
\end{description}
Here $x_1$ is the distance between two cars at time $t=110$ in Case 3 and that when collision occurs in Case 4. Moreover, $c$ is set to -0.01. As seen in Case 1, $r_1$ does not give any reward even if controlling succeeds. On the other hand, the sooner the following car enters the range $[l_1, l_2]$, the larger positive reward $r_2$ gives. $r_1$ is defined as a penalty to prevent failure. $r_2$ includes not only penalties but also rewards to complete the goal as soon as possible.

Steps in the learning algorithm are as following:
\begin{description}
\vspace{2mm}
  \item[Step 1 (Learning phase):]The  system controls a car in a learning problem using stochastic policies $\pi$ in Eq. \eqref{eq:5} or $\pi'$ in Eq. \eqref{eq:8}. Rule weights are updated according to learning rules at the end of the episode. Rule weights are normalized as $\sum_i \theta_i=1.0$i$\theta_i\geq 0$j. These procedures are repeated sequentially to all learning problems from Problem 1 to Problem 16.
  \item[Step 2 (Evaluation phase):]The control system controls a car in a learning problem by deterministic outputs $y_{\textrm{G}}(t)$ in Eq. \eqref{eq:6} or $y_{\textrm{O}}(t)$ in Eq. \eqref{eq:14} and then checks whether Requisites 1 and 2 are satisfied in that episode. This procedure is repeated sequentially for all learning problems from Problem 1 to Problem 16.
  \item[Step 3 (Termination conditions):]If the goals are completed in all learning problems at Step 2, finish learning. Otherwise go to Step 1 unless the number of times repeating Step 1 and Step 2 gets larger than 200. The repeating number $m_{\textrm{c}}$ is called the learning number of times.
\vspace{2mm}
\end{description}
The initial values of parameters are set as $\theta_i=0.05$ for all $i$, $T=T'=0.04$, $\varepsilon=0.0075$ and $\varepsilon '=0.0003$. Each learning experiment consists of processes from Step 1 to Step 3, and the learning system outputs a set of values of $\{\theta_i\}\ (i=1,\cdots,20)$ at the end of Step 3. We repeated the learning experiments 10,000 times and obtained 10,000 sets of weights.

\subsection{Results of learning experiments and discussion}

Let $S$ be a set of weight vectors $\mbox{\boldmath $s$}=(\theta_1,\cdots,\theta_{20})$ that are obtained by the learning and that satisfy the two requisites of all 16 problems in Table \ref{table:2}. We define the following two simplified conditions that check whether $s$ causes rapid changes in speed while controlling a car.
\begin{description}
\vspace{2mm}
  \item[Condition 1:]The difference between values of the following car's speed at $t=109$ and $t=110$, which is equal to the absolute value of acceleration at $t=109$, is smaller than 0.1 [km/h]. 
  \item[Condition 2:]The difference in speed between the two cars is smaller than 0.1 [km/h] at $t=110$.
\vspace{2mm}
\end{description}

The results of the learning experiments are shown in Table \ref{table:3}. A subset of $S$ is denoted as $S_{\textrm{c}}$ of which element $s$ brings episodes satisfying the two conditions. Therefore, $S_{\textrm{c}}$ is a set of good solutions without rapid change or oscillation in speed. The average of $m_{\textrm{c}}$'s and that of $t_{\textrm{in}_{\textrm{c}}}$'s are denoted as $\overline{m_{\textrm{c}}}$ and $\overline{t_{\textrm{in}_{\textrm{c}}}}$ respectively. $t_{\textrm{in}_{\textrm{c}}}$ is $t_{\textrm{in}}$ in the case of episodes produced by rule sets in $S_{\textrm{c}}$.

\renewcommand{\arraystretch}{1.5}
\begin{table}[!tb]
\caption{Results of learning experiments after each learning method is repeated 10,000 times.}
\label{table:3}
\begin{center}
\begin{tabular}{c|c|c|c|c|c|c|c|c}
\noalign{\hrule height 0.4mm}
\hline
\multirow{2}{*}{Meth.} & \multicolumn{4}{|c}{$r=r_1$} & \multicolumn{4}{|c}{$r=r_2$} \\
\cline{2-9}
 & $|S|$ & $|S_{\textrm{c}}|$ & $\overline{m_{\textrm{c}}}$ & $\overline{t_{\textrm{in}_{\textrm{c}}}}$ & $|S|$ & $|S_{\textrm{c}}|$ & $\overline{m_{\textrm{c}}}$ & $\overline{t_{\textrm{in}_{\textrm{c}}}}$\\
\hline
$\textrm{(i)}\ $ & 9873 & 9036 & 25.1 & 17.6 & 9890 & 8673 & 19.3 & 17.4\\
\hline
$\textrm{(i\hspace{-.08em}i)}\ $ & 9921 & 9087 & 36.8 & 16.8 & 9975 & 9040 & 16.7 & 16.5\\
\hline
$\textrm{(i\hspace{-.08em}i\hspace{-.08em}i)}\ $ & 9908 & \cellcolor[gray]{0.85} 9539 & 36.4 & 17.0 & 9951 & \cellcolor[gray]{0.85} 9686 & 18.8 & 17.2\\
\hline
\noalign{\hrule height 0.4mm}
\end{tabular}
\end{center}
\end{table}
\renewcommand{\arraystretch}{1.0}

The values of $|S|$ in Table \ref{table:3} show that feasible solutions were obtained in more than $98\%$ of the learning experiments using any method from (i) to (iii) in Table \ref{table:5}. According to the values of $|S_{\textrm{c}}|$, good solutions that do not cause rapid change and oscillation in speed were obtained frequently. The size of set $S$ in (ii) and (iii) has increased over (i) in the cases of both $r_1$ and $r_2$ (see grey cells in Table \ref{table:3}). That means the improvements introduced in 4.2 are effective in suppressing undesirable vibration.

Next, we compare the effect of $r_1$ with that of $r_2$. Looking at values of $\overline{m_{\textrm{c}}}$ in Table \ref{table:3}, $r_2$ is more effective in reducing the learning number of times than $r_1$. The effect of reducing $\overline{t_{\textrm{in}_{\textrm{c}}}}$ by $r_2$, however, is not so clear. Results of $|S_{\textrm{c}}|$ indicate that $r_2$ reduces the number of good solutions and gives worse results than $r_1$ in some cases. 

\subsection{Control problems for evaluating rule sets}

We prepared 697 control problems\footnote{We tested and found that the system cannot control the car well in these test problems if all $\theta_i$ are fixed at 0.05 by preliminary experiments beforehand.} for evaluating the rule weights obtained by learning in 6.1. The test problems consists of two datasets. The first dataset has $5\times5\times5\times5=625$ records. Conditions of each of them are a leading/following car's initial speed $\subseteq\{45,55,65,75,85\}/\{0,10,30,50,70\}$ [km/h], a initial distance between the two cars $\subseteq\{30,45,65,70,80\}$ [m], and a target distance $\subseteq\{[10,30], [20,40], [40,60], [50,60], [60,70]\}$ [m]. The second dataset consists of a leading/following car's initial speed $\subseteq\{40,50,60\}/\{20,40,60\}$ [km/h], an initial distance between the two cars $\subseteq\{20,40,60\}$ [m], and a target distance $\subseteq\{[10,20], [40,50], [45,60]\}$ [m]. The cases when the leading car's speed was 50 [km/h] and the initial distance was 20 [m] were excluded in the evaluation experiments. Therefore, the second dataset has $3\times3\times3\times3-3\times3=72$ records. All rule weights $s\in S_{\textrm{c}}$ were tested by the procedure in the evaluation phase at Step 2 in 6.1 using the 697 test control problems.

\subsection{Results of evaluation experiments and discussion}

\renewcommand{\arraystretch}{1.5}
\begin{table}[!tb]
\caption{Results of experiments to evaluate values of $s\in S_{\textrm{c}}$ obtained by learning.}
\label{table:4}
\begin{center}
\begin{tabular}{c|c|c|c|c|c|c|c|c}
\noalign{\hrule height 0.4mm}
\hline
\multirow{2}{*}{Meth.} & \multicolumn{4}{|c}{$r=r_1$} & \multicolumn{4}{|c}{$r=r_2$} \\
\cline{2-9}
 & $|S'|$ & $|S'_{\textrm{c}}|$ & $|S'_{\textrm{c}}|/|S_{\textrm{c}}|$ & $\overline{t'_{\textrm{in}_{\textrm{c}}}}$ & $|S'|$ & $|S'_{\textrm{c}}|$ & $|S'_{\textrm{c}}|/|S_{\textrm{c}}|$ & $\overline{t'_{\textrm{in}_{\textrm{c}}}}$\rule[0pt]{0pt}{1pt}\\
\hline
$\textrm{(i)}\ $ & 569 & 563 & 0.062 & 19.5 & 545 & 543 & 0.063 & 19.3\\
\hline
$\textrm{(i\hspace{-.08em}i)}\ $ & 670 & 668 & 0.074 & 20.0 & 576 & 573 & 0.063 & 19.6\\
\hline
$\textrm{(i\hspace{-.08em}i\hspace{-.08em}i)}\ $ & 761 & 760 & 0.080 & 19.9 & 649 & 649 & 0.067 & 19.5\\
\hline
\noalign{\hrule height 0.4mm}
\end{tabular}
\end{center}
\end{table}
\renewcommand{\arraystretch}{1.0}

All sets of rule weights in $S_{\textrm{c}}$ are tested by evaluation experiments in 6.3. We define subset $S'\ (\subseteq S_{\textrm{c}})$ by a set of rule weights that completed the goals in all of the test problems. $S'_{\textrm{c}}$ is defined by a set of $s$ that are members of $S'$ and produced only episodes compatible with both Conditions 1 and 2 in 6.2. Therefore, it holds that $S'_{\textrm{c}}\subseteq S'\subseteq S_{\textrm{c}}$. Results of evaluation experiments are shown in Table \ref{table:4}. $\overline{t'_{\textrm{in}_{\textrm{c}}}}$ is the average of time step $t'_{\textrm{in}_{\textrm{c}}}$ when $s\in S'_{\textrm{c}}$ were used.

Values of $|S'_{\textrm{c}}|$ in Table \ref{table:4} indicate that good control without oscillation is possible in the test problems. The number becomes gradually large in order of (i), (ii) and (iii) in the cases of both $r_1$ and $r_2$ as observed in learning experiments in 6.2. The ratio $|S'_{\textrm{c}}|/|S_{\textrm{c}}|$ means how many of the good solutions in the learning problems are also good in the test problems, i.e., generality of solutions obtained by learning. The ratio gradually grows in the same order. These results mean the objective function $E'(y)$ in Eq. \eqref{eq:7} is effective in suppressing undesirable vibration even in the test problems.

Next, we compare the effect of $r_1$ with that of $r_2$ to the test problems. The values of $\overline{t'_{\textrm{in}_{\textrm{c}}}}$ in Table \ref{table:4} show that $r_2$'s effect in reducing the time until entering the target range of distance is not so large, and $r_2$ reduces the number of good solutions. The latter is one of unanticipated side effects caused by $r_2$. If reward functions such as $r_2$ are used to shorten the time until entering the range of distance, the objective of learning changes to another one. Reward functions should be designed not to cause such unanticipated side effects.

\subsection{Examples of rule weight parameters and speed control after learning}

\begin{figure}[tb]
\centering
\begin{center}
\includegraphics[bb=0 0 420 139, scale=1]{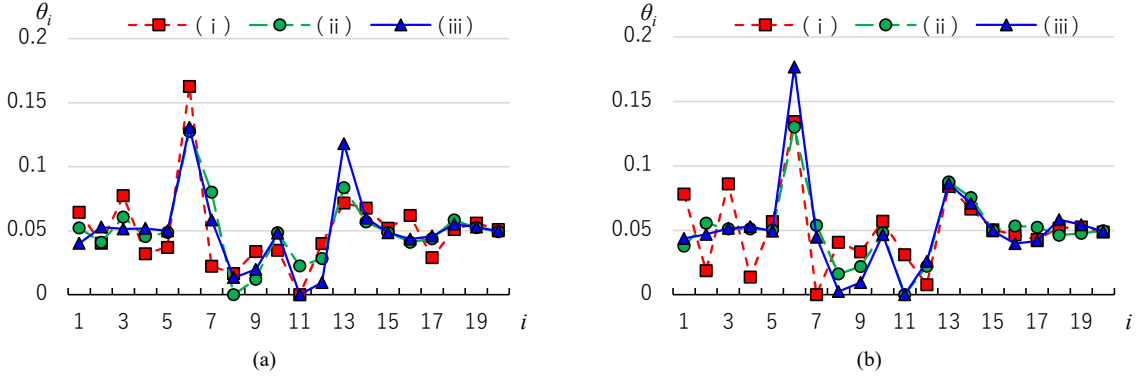}
\end{center}
\vspace{-5mm}
\caption{Set of rule weights. (a) Set of rule weights learned with reward function $r_1$. (b) Set of rule weights learned with reward function $r_2$.}
\label{fig:2}
\vspace{5mm}
\end{figure}

\begin{figure}[tb]
\centering
\begin{center}
\includegraphics[bb=0 0 434 143, scale=1]{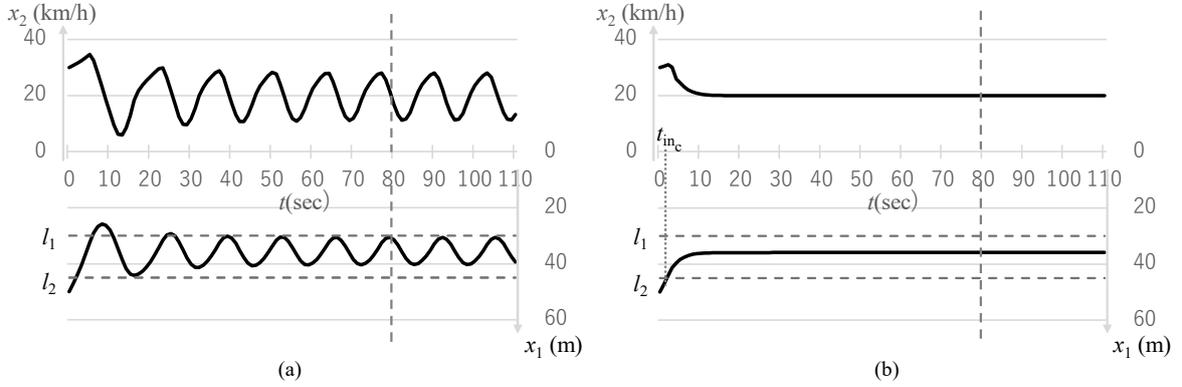}
\end{center}
\vspace{-5mm}
\caption{Results of the speed control. (a) Results of the speed control with a weight set $s\notin S_{\textrm{c}}$ learned by Method (i) and reward function $r_2$. (b) Results of the speed control with a weight set $s\in S_{\textrm{c}}$ learned by Method (i) and reward function $r_2$.}
\label{fig:3}
\end{figure}

Figure \ref{fig:2} shows the rule weight values of $s\in S'_{\textrm{c}}$ that clear all of the test problems and minimize $\overline{t'_{\textrm{in}_{\textrm{c}}}}$ observed in the evaluation experiments in 6.4. In Fig. \ref{fig:2}, the weights of Rule 6 and Rule 13 become larger than other rules after learning, while Rule 8 and Rule 11 become smaller than other rules. Rule 6 says that strong acceleration is necessary if the distance to the front car is long and the following car's speed is slow as seen in Table 1. This rule seems to be quite correct. Rule 13 is also a correct rule because it says that strong deceleration is necessary if the following distance is short and the following car's speed is fast. By contrast, Rule 8 and Rule 11 have the exact opposite meaning to Rule 6 and Rule 13. Rule 8 and Rule 11 are not correct and must be suppressed. This reinforcement of Rules 6 \& 13 and suppression of Rules 8 \& 11 as seen in Fig. \ref{fig:2} are very effective and desirable in controlling a car's speed independent of learning methods from (i) to (iii) and reward functions $r_1$ and $r_2$.

The meanings of Rules 1-5 and 16-20 are not so clear. They do not seem to have important roles in controlling a car's speed. Figure \ref{fig:2} shows that their weights did not change from their initial values ($=0.05$) after learning by Methods (ii) and (iii). However, the values changed very much after learning by Method (i). This difference comes from the existence of regularization functions in the three learning methods. The gravity-center model and the time-smoothing constraint in objective function (11) of Methods (ii) and (iii) are effective in reducing unnecessary change in the weights of unimportant rules.

Figure \ref{fig:3} shows the change in the following distance and the following car's speed in Problem 1 in Table 3. Figure \ref{fig:3}(a) is an example of rule weight $s$ that does not belong to $S_{\textrm{c}}$ and caused oscillation. Figure \ref{fig:3}(b) is an example of $s\in S_{\textrm{c}}$. Rule weight $s$ in both the cases was learned by Method (i) with reward $r_2$. Rule weight $s\notin S_{\textrm{c}}$, which produces a trajectory as shown in Fig. \ref{fig:3}(a), is certainly a solution that reaches the goals of controlling, but it is not a desirable solution from the viewpoints of passenger comfort and ease of control.

\section{Conclusion} \label{sec7}

A fuzzy control system with parameters learned by policy gradient reinforcement learning was previously proposed by Igarashi and Ishihara \cite{Igarashi2}. That system makes an action decision and deals with continuous-value output as an agent's action. For this purpose, the expectations generated by a stochastic action-selection policy, which is called a gravity-center model, were used to determine the control output. This paper tried to improve the control and learning method by selecting an output near the gravity-center and introducing a time-smoothing constraint term in the objective function used in the stochastic policy to avoid rapid changes in output. This paper derived the new learning rule for rule weights from the new objective function.

This improved control with the new learning method was applied to the speed-control problems of an automobile in maintaining an appointed distance from the car moving in front of it during a specific time interval. The results of learning experiments with fuzzy-rule weights and evaluation experiments showed the effect of the proposed improvements in reducing undesirable rapid change and oscillation in speed. Two types of reward functions were proposed and tested in these learning experiments. One was designed to give punishment when the goal of the control was not reached but no reward even if the control succeeded. The other reward function was designed to reach the goals as soon as possible. The latter greedy reward function reduced the number of learning times necessary to reach the goal. However, that one is apt to produce undesirable oscillation in speed more easily than the former type. Accordingly, reward functions must be designed so as not to produce any side effect. 

In the future, we will consider appropriate reward functions and learning of parameters included in the membership functions of fuzzy control rules.


\appendix
\section{Proof of Eq. \eqref{eq:10}}

If $\pi'\left( y(t)\right)$ is a function in Eq. \eqref{eq:8}, characteristic eligibility $e'_{\theta}(t)$ can be written \cite{Igarashi1} as
\begin{eqnarray}
e'_{\theta}(t) &\equiv& \frac{\partial}{\partial \theta}\textrm{ln}\pi'\left( y(t)\right) \nonumber\\
&=& -\frac{1}{T'}\left\{\frac{\partial E'(y(t))}{\partial \theta}-\left\langle\frac{\partial E'(y)}{\partial \theta}\right\rangle_{\pi'(y;x(t))}\right\}.
\label{eq:A1}
\end{eqnarray}
The first term in the braces \{c\} in Eq. \eqref{eq:A1} can be written as
\begin{eqnarray}
\frac{\partial E'\left(y(t)\right)}{\partial \theta} &=& \left( y(t)-y_{\textrm{G}}(t)\right) \frac{\partial y_{\textrm{G}}(t)}{\partial \theta}
\label{eq:A2}
\end{eqnarray}
if $E'(y(t))$ in Eq. \eqref{eq:7} is substituted to Eq. \eqref{eq:A1}. The last term in Eq. \eqref{eq:A2} is expressed as
\begin{eqnarray}
\frac{\partial y_{\textrm{G}}(t)}{\partial \theta} &=& \frac{\partial \left\langle y(t)\right\rangle_{\pi(y;x(t))}}{\partial \theta}\nonumber\\
&=& \sum_{y}y\frac{\partial}{\partial \theta}\pi(y;x(t),\theta)\nonumber\\
&=& \sum_{y}\pi(y;x(t),\theta)y\frac{\partial}{\partial \theta}\textrm{ln}\pi(y;x(t),\theta)\nonumber\\
&=& \left\langle ye_{\theta}(t)\right\rangle_{\pi(y;x(t))},
\label{eq:A3}
\end{eqnarray}
where
\begin{eqnarray}
e_{\theta}(t) &\equiv& \frac{\partial}{\partial \theta}\textrm{ln}\pi(y(t))\nonumber\\
&=& -\frac{1}{T}\left\{\frac{\partial E(y(t))}{\partial \theta}-\left\langle\frac{\partial E(y)}{\partial \theta}\right\rangle_{\pi(y;x(t))}\right\}.
\label{eq:A4}
\end{eqnarray}
Substituting Eq. \eqref{eq:A4} into Eq. \eqref{eq:A3}, the first term in Eq. \eqref{eq:A1} becomes
\begin{eqnarray}
\frac{\partial y_{\textrm{G}}(t)}{\partial \theta} &=& -\frac{1}{T}\left\langle{ y\left\{\frac{\partial E(y)}{\partial \theta}-\left\langle\frac{\partial E(y)}{\partial \theta}\right\rangle_{\pi(y;x(t))}\right\}}\right\rangle_{\pi(y;x(t))}\nonumber\\
&=& -\frac{1}{T}\left\{ \left\langle y\frac{\partial E(y)}{\partial \theta}\right\rangle_{\pi(y;x(t))}-\left\langle y\right\rangle_{\pi(y;x(t))}\left\langle\frac{\partial E(y)}{\partial \theta}\right\rangle_{\pi(y;x(t))}\right\}\nonumber\\
&=& -\frac{1}{T}\left\langle(y-y_{\textrm{G}}(t))\frac{\partial E(y)}{\partial \theta}\right\rangle_{\pi(y;x(t))}.
\label{eq:A5}
\end{eqnarray}
Next, let us consider the second term in Eq. \eqref{eq:A1}. We calculate the expectation of Eq. \eqref{eq:A2} and get
\begin{eqnarray}
\left\langle\frac{\partial E'(y)}{\partial \theta}\right\rangle_{\pi'(y;x(t))} &=& \left\langle(y-y_{\textrm{G}}(t))\frac{\partial y_{\textrm{G}}(t)}{\partial \theta}\right\rangle_{\pi'(y;x(t))}\nonumber\\
&=& \left\langle y\frac{\partial y_{\textrm{G}}(t)}{\partial \theta}\right\rangle_{\pi'(y;x(t))}-\left\langle y_{\textrm{G}}(t)\frac{\partial y_{\textrm{G}}(t)}{\partial \theta}\right\rangle_{\pi'(y;x(t))}\nonumber\\
&=& \left(\left\langle y\right\rangle_{\pi'(y;x(t))}-y_{\textrm{G}}(t)\right)\frac{\partial y_{\textrm{G}}(t)}{\partial \theta}.
\label{eq:A6}
\end{eqnarray}
Substituting Eqs. \eqref{eq:A2} and \eqref{eq:A6} into Eq. \eqref{eq:A1}, we get
\begin{eqnarray}
e'_{\theta}(t) &=& -\frac{1}{T'}\Biggl\{\left( y(t)-y_{\textrm{G}}(t)\right) \frac{\partial y_{\textrm{G}}(t)}{\partial \theta}-\left(\left\langle y\right\rangle_{\pi'(y;x(t))}-y_{\textrm{G}}(t)\right)\frac{\partial y_{\textrm{G}}(t)}{\partial \theta}\Biggr\}\nonumber\\
&=& -\frac{1}{T'}\frac{\partial y_{\textrm{G}}(t)}{\partial \theta}\left(y(t)-\left\langle y\right\rangle_{\pi'(y;x(t))}\right).
\label{eq:A7}
\end{eqnarray}
Using Eq. \eqref{eq:A5}, this equation becomes
\begin{eqnarray}
e'_{\theta}(t) &=& \frac{1}{TT'}\left\langle(y-y_{\textrm{G}}(t))\frac{\partial E(y)}{\partial \theta}\right\rangle_{\pi(y;x(t))}\left(y(t)-\left\langle y\right\rangle_{\pi'(y;x(t))}\right).
\label{eq:A8}
\end{eqnarray}
According to Eq. \eqref{eq:2}, ${\partial E(y)}/{\partial \theta}$ is easily calculated and expressed in a very simple form as
\begin{eqnarray}
\frac{\partial E(y;x(t),\theta,A,B)}{\partial \theta_i} &=& \frac{\partial }{\partial \theta_i}\left[-\sum_{i=1}^{n_R}\theta_i A^{i}(x(t))B^{i}(y)\right]\nonumber\\
&=& -A^{i}(x(t))B^{i}(y).
\label{eq:A9}
\end{eqnarray}
If Eq. \eqref{eq:A9} is substituted into Eq. \eqref{eq:A8}, Eq. \eqref{eq:10} is obtained.
\begin{eqnarray}
e'_{\theta_i}(t) &=& -\frac{1}{TT'}\left\langle(y-y_{\textrm{G}}(t))A^{i}(x(t))B^{i}(y)\right\rangle_{\pi(y;x(t))}\left(y(t)-\left\langle y\right\rangle_{\pi'(y;x(t))}\right)\nonumber\\
&=& -\frac{A^{i}(x(t))}{TT'}\left\langle(y-y_{\textrm{G}}(t))B^{i}(y)\right\rangle_{\pi(y;x(t))}\left(y(t)-\left\langle y\right\rangle_{\pi'(y;x(t))}\right).
\label{eq:A10}
\end{eqnarray}

\bibliographystyle{model1-num-names}

\end{document}